\documentclass{ecai}
\usepackage{latexsym}
\usepackage{amssymb}
\usepackage{amsmath}
\usepackage{amsthm}
\usepackage{booktabs}
\usepackage{enumitem}
\usepackage{graphicx}
\usepackage{color}
\usepackage[utf8]{inputenc}
\usepackage{array}     % For advanced column specifications in tables
\usepackage{booktabs}    % For professional quality tables

\newcommand{\BibTeX}{B\kern-.05em{\sc i\kern-.025em b}\kern-.08em\TeX}

\begin{document}

%%%%%%%%%%%%%%%%%%%%%%%%%%%%%%%%%%%%%%%%%%%%%%%%%%%%%%%%%%%%%%%%%%%%%%%%

\begin{frontmatter}

\paperid{63}

%%% Use this command to specify the title of your paper.
\title{Agentic RAG with Knowledge Graphs for Complex Multi-Hop Reasoning in Real-World Applications}

%%% Use this combinations of commands to specify all authors of your
%%% paper. Use \fnms{} and \snm{} to indicate everyone's first names
%%% and surname. This will help the publisher with indexing the
%%% proceedings. Please use a reasonable approximation in case your
%%% name does not neatly split into "first names" and "surname".
%%% Specifying your ORCID digital identifier is optional.
%%% Use the \thanks{} command to indicate one or more corresponding
%%% authors and their email address(es). If so desired, you can specify
%%% author contributions using the \footnote{} command.

\author[A]{\fnms{Jean}~\snm{Lelong}}
\author[A]{\fnms{Adnane}~\snm{Errazine}}
\author[A]{\fnms{Annabelle}~\snm{Blangero}}

\address[A]{Ekimetrics, France}
%\address[B]{INRAE}
%\address[C]{Short Alternate Affiliation of Third Author}

%%% Use this environment to include an abstract of your paper.

\begin{abstract}
Conventional Retrieval-Augmented Generation (RAG) systems enhance Large Language Models (LLMs) but often fall short on complex queries, delivering limited, extractive answers and struggling with multiple targeted retrievals or navigating intricate entity relationships. This is a critical gap in knowledge-intensive domains. We introduce INRAExplorer, an agentic RAG system for exploring the scientific data of INRAE (France's National Research Institute for Agriculture, Food and Environment). INRAExplorer employs an LLM-based agent with a multi-tool architecture to dynamically engage a rich knowledge base, through a comprehensive knowledge graph derived from open access INRAE publications. This design empowers INRAExplorer to conduct iterative, targeted queries, retrieve exhaustive datasets (e.g., all publications by an author), perform multi-hop reasoning, and deliver structured, comprehensive answers. INRAExplorer serves as a concrete illustration of enhancing knowledge interaction in specialized fields.
\end{abstract}

\end{frontmatter}

%%%%%%%%%%%%%%%%%%%%%%%%%%%%%%%%%%%%%%%%%%%%%%%%%%%%%%%%%%%%%%%%%%%%%%%%
\section{Introduction}

Effective digital knowledge utilization demands relevant, exhaustive and structured information retrieval. While Retrieval-Augmented Generation (RAG) grounds Large Language Models (LLMs) in curated, trustworthy information \citep{lewis2020retrieval}, prevalent architectures —termed classical RAG— exhibit key limitations. Classical RAG typically retrieves a limited set (top-k) of semantically similar text chunks via vector search \citep{karpukhin2020dense, gunther2023jina} to contextualize an LLM. While valuable for anchoring LLM responses and extractive question answering, this 'top-k snippet' approach is often insufficient for queries requiring exhaustive lists, synthesis from multiple distinct data points, or navigating complex relational paths (e.g., author to publications to funding projects).

To address these shortcomings, we introduce INRAExplorer, a method that synergizes agentic RAG \citep{singh2025agenticsurvey} for dynamic reasoning and Knowledge Graph (KG)-enhanced RAG \citep{zhang2025surveygraphrag, zhu2025knowledgegraphguidedrag} for structured, exhaustive retrieval. INRAExplorer deeply incorporates KG querying as a core agentic capability, enabling it to overcome the single-pass, limited-context nature of classical RAG. This fusion delivers precise, relationally-aware retrieval from KGs, combined with the adaptive, multi-hop reasoning of an LLM-driven agent.

While the integration of Knowledge Graphs with LLMs is gaining traction, many current approaches primarily use KGs by performing a sophisticated form of map-reduce summarization over graph-retrieved data \citep{edge2024graphrag}. In contrast, INRAExplorer, drawing inspiration from the need for deeper reasoning similar to human investigative processes, focuses on enabling the LLM agent to construct chains of thought leveraging several ways to retrieve information. Our system empowers the agent to dynamically navigate between different tools, gathering evidence, evaluating intermediate findings, and planning subsequent steps. This allows INRAExplorer to act more like a human researcher, meticulously assembling pieces of information to construct a comprehensive and nuanced answer, rather than simply summarizing pre-existing snippets of information.

%%%%%%%%%%%%%%%%%%%%%%%%%%%%%%%%%%%%%%%%%%%%%%%%%%%%%%%%%%%%%%%%%%%%%%%%
% Section 2: The INRAExplorer Method: Agentic RAG with Knowledge Graphs
%%%%%%%%%%%%%%%%%%%%%%%%%%%%%%%%%%%%%%%%%%%%%%%%%%%%%%%%%%%%%%%%%%%%%%%%

\begin{table*}[t] % Use [t] to place it at the top of a page, spans both columns.
    \centering
    \caption{Distribution of Node Types in the INRAExplorer Knowledge Graph (Total Nodes: 417,030)}
    \label{tab:kg_nodes} % Retain the original label for the table
    % The 'p{10cm}' for the last column allows text wrapping for the descriptions.
    % Adjust the width (10cm) as needed depending on your document's specific column width and font size.
    \begin{tabular}{@{}lrrp{10cm}@{}}
    \toprule
    Node Type & Count & Percentage & Description \\
    \midrule
    Author & 233,728 & 56.0\% & Researchers and authors of scientific publications \\
    Keyword & 96,588 & 23.2\% & Keywords associated with publications (declared by authors) \\
    Publication & 38,791 & 9.3\% & Scientific articles and other academic publications \\
    Software & 21,617 & 5.2\% & Software developed or used in research \\
    Concept & 13,591 & 3.3\% & Concepts from the INRAE thesaurus identified in publications \\
    Journal & 5,563 & 1.3\% & Scientific journals where works are published \\
    Project & 3,999 & 1.0\% & Funded research projects \\
    Domain & 2,595 & 0.6\% & Thematic domains of the INRAE thesaurus \\
    ResearchUnit & 299 & 0.1\% & INRAE research units and laboratories \\
    Dataset & 240 & 0.1\% & Datasets used or produced in research \\
    Region & 19 & 0.0\% & Geographic regions where research units are located \\
    \bottomrule
    \end{tabular}
\end{table*}

\section{INRAExplorer: Agentic RAG with Knowledge Graphs}

INRAExplorer is a system designed to implement and showcase the capabilities of this agentic, KG-enhanced RAG methodology. It employs an LLM-based agent that orchestrates a suite of specialized tools to interact with a hybrid knowledge base, specifically tailored to the complexities of real-world scientific data exploration.

\subsection{Knowledge Base Construction}

The foundation of INRAExplorer is a comprehensive knowledge base constructed from INRAE's scientific output, focusing on publications from January 2019 to August 2024 and restricted to Open Access documents.
\textbf{Core data} originates from an internal join of publication metadata from HAL (Hyper Articles en Ligne), valued for its qualitative and exhaustive INRAE scope, and OpenAire, used for deduplication and enriched author/project information. This merged dataset was further \textbf{enriched} with other public sources using DOIs, including BBI (Base Bibliographique INRAE) for validation, ScanR for additional publication and project details, and dataset repositories for links to underlying research data.

Full-text content of Open Access PDF publications was processed using GROBID for structured text extraction \citep{lopez2009grobid}, isolating sections such as title, abstract, keywords, introduction, and conclusion to form meaningful chunks.

A significant component of the structured knowledge comes from the \textbf{INRAE Thesaurus}. This was integrated into the KG as a dedicated subgraph, where hierarchical terms form `Domain` nodes and more specific, leaf-level terms are represented as `Concept` nodes (see Table~\ref{tab:kg_nodes}). Publications are then linked to these `Concept` nodes using exact matches in selected sections of publication texts, providing a controlled vocabulary for semantic exploration and enabling the model to reference the thesaurus for understanding domain-specific query vocabulary.

The processed information is stored in a \textbf{hybrid knowledge base}:
\begin{itemize}
    \item A \textbf{Vector Database} stores representations of the textual chunks. For each publication, key sections (title, abstract, introduction, conclusion) were concatenated to form these chunks. Two types of vectors were computed for each chunk to support hybrid search: a dense vector using a Jina v3 embedding model for semantic similarity \citep{gunther2023jina}, and a sparse vector using BM25 for keyword-based matching \citep{robertson2009probabilisticbm25}.
    \item A \textbf{Knowledge Graph} models the core entities and their multifaceted relationships \citep{Roll2025Augmenting}. The graph comprises various node types—including `Publication`, `Author`, `Keyword`, but also `Concept` and `Domain` from INRAE's thesaurus, and several other complementary meta-information. Table~\ref{tab:kg_nodes} details these node types and their distributions. The graph contains more than 1 million relationships that link these entities. This KG serves as the backbone for precise, structured queries, multi-hop reasoning \citep{edge2024graphrag, he2024gretriever}, and the retrieval of exhaustive, interconnected information sets.
\end{itemize}

\subsection{Agent-Driven Multi-Tool Orchestration}

At the heart of INRAExplorer is an LLM-based agent utilizing the open-weight model deepseek-r1-0528. Its primary responsibilities include: understanding the user's query, decomposing it if necessary, formulating a plan of action, dynamically selecting and invoking the appropriate tools, and synthesizing the information gathered from multiple tool calls into a coherent and comprehensive response.

The agent has access to a set of specialized tools, each designed for a specific type of interaction with the knowledge base:
\begin{itemize}
    \item \textbf{\texttt{SearchGraph} (Knowledge Graph Querying):} This is the main tool, pivotal for deep interaction with the Neo4j KG. It allows the agent to send Cypher queries to retrieve specific entities, traverse complex relationship paths, and gather exhaustive lists (e.g., all publications by a specific author, all projects associated with a particular research unit). This tool is critical for obtaining structured and complete answers that go beyond simple text snippet retrieval.
    \item \textbf{\texttt{SearchPublications} (Hybrid Publication Search):} This helper tool interfaces with the vector database to perform hybrid searches (semantic \citep{karpukhin2020dense, gunther2023jina} and keyword-based \citep{robertson2009probabilisticbm25}) over the corpus of publication texts. It allows the agent to find relevant entry points into the graph to start its reasoning and chain of retrieval, for instance by identifying initial sets of documents for further exploration via the KG. A reranker (e.g., Cohere) is used to refine the top results and merge the results of the two retrieval methods.
    \item \textbf{\texttt{SearchConceptsKeywords} (Concept/Keyword Search):} This other helper tool allows the agent to find relevant entry points into the graph. It helps bridge the gap between user queries and the structured vocabulary of the knowledge base by allowing the agent to search for relevant concepts from the integrated thesauri or the author keywords indexed in the KG. This is useful for query disambiguation, suggesting related terms, or finding precise entry points for subsequent graph traversal.
    \item \textbf{\texttt{IdentifyExperts} (Expert Identification):} This refined tool encapsulates complex domain-specific knowledge for identifying experts on a given topic. It allows for more reproducibility through different answers by the model. The tool uses \texttt{SearchPublications} to find highly relevant papers, then use \texttt{SearchGraph} to analyze the authors of these papers, their number of citations on this given topic, their collaboration networks, their involvement in related projects, and other structural indicators of expertise, before synthesizing a ranked list based on a composite score.
\end{itemize}

%%%%%%%%%%%%%%%%%%%%%%%%%%%%%%%%%%%%%%%%%%%%%%%%%%%%%%%%%%%%%%%%%%%%%%%%
% Section 3: Illustrative Scenarios and Capabilities
%%%%%%%%%%%%%%%%%%%%%%%%%%%%%%%%%%%%%%%%%%%%%%%%%%%%%%%%%%%%%%%%%%%%%%%%

\section{Illustrative Scenarios and Capabilities}

The INRAExplorer method, through its agentic and Knowledge Graph (KG)-centric design, offers significant advantages over classical RAG systems, particularly for complex information needs requiring exhaustive and structured answers. This section illustrates these capabilities through two representative scenarios, highlighting multi-hop sequential reasoning and the use of specialized, modular tools.

\subsection{Multi-Hop Sequential Reasoning for Complex Queries}

Many real-world queries necessitate navigating multiple types of relationships across different entities, a task where classical RAG often falters. For example, consider the query: "Find INRAE authors who have published on 'climate change adaptation strategies', identify the projects that funded these publications, and list other key topics these funding projects are related to."

INRAExplorer's agent tackles this through a sequence of reasoned steps, demonstrating multi-hop reasoning:
\begin{enumerate}
    \item \textbf{Initial Information Gathering}: The agent first uses \texttt{SearchPublications} or \texttt{SearchConceptsKeywords} to identify an initial set of relevant publications and their authors related to 'climate change adaptation strategies'. This step grounds the query in the available literature.
    \item \textbf{First Hop - Identifying Funding}: Using the \texttt{SearchGraph} tool, the agent then queries the KG to find `Project` nodes linked to these initial publications via a `FUNDED\_BY` relationship. This constitutes the first "hop" in the reasoning chain, connecting publications to their funding sources.
    \item \textbf{Second Hop - Exploring Related Project Topics}: Subsequently, for each identified funding project, the agent again employs \texttt{SearchGraph}. This time, it seeks other `Concept` nodes (representing topics) linked to these projects via relationships like `DESCRIBES`. This is the second "hop," broadening the understanding of the projects' thematic scope, through common publication nodes.
    \item \textbf{Synthesis}: Finally, the agent synthesizes these interconnected findings into a structured answer. This response explicitly shows the chain: authors $\rightarrow$ publications on 'climate change adaptation strategies' $\rightarrow$ funding projects $\rightarrow$ other related research topics.
\end{enumerate}
This process illustrates how the agent can meticulously assemble pieces of information, retrieve complete sets of entities and their specific relationships from the KG, and construct a comprehensive answer that goes far beyond simple snippet retrieval.

\subsection{Modular and Controlled Expertise Identification}

INRAExplorer's architecture supports the integration of specialized, high-level tools that encapsulate complex, domain-specific logic, offering more controlled and reproducible outputs compared to relying solely on raw tool access for every task. The \texttt{IdentifyExperts} tool is a prime example of this modularity.

Consider a query such as: "Identify leading INRAE experts on 'zoonoses'."
Instead of the agent needing to devise a multi-step plan from scratch using basic tools, it can leverage the \texttt{IdentifyExperts} tool. This tool is designed to provide a replicable and controlled method for expert identification:
\begin{enumerate}
    \item \textbf{Tool Invocation}: The agent recognizes the query's intent and calls the \texttt{IdentifyExperts} tool with the topic 'zoonoses'.
    \item \textbf{Encapsulated Workflow}: The \texttt{IdentifyExperts} tool executes a predefined sequence of actions, which itself involves calls to other foundational tools:
    \begin{itemize}
        \item It first uses \texttt{SearchPublications} to find highly relevant papers on 'zoonoses'.
        \item Then, it employs \texttt{SearchGraph} to extract the authors of these papers.
        \item For each author, it calculates a composite expertise score based on multiple weighted metrics: average relevance of their articles to the topic, number of articles in the top 10\% of results, total number of relevant publications, citation counts for these publications, period of activity in the domain, and recency of their latest publication.
    \end{itemize}
    \item \textbf{Structured Output}: The tool returns a ranked list of experts on 'zoonoses', along with their expertise scores and the breakdown of these scores.
    \item \textbf{Synthesis by Agent}: The agent then presents this structured information to the user.
\end{enumerate}

This approach ensures consistent, domain-aware execution for tasks like expert identification and simplifies agent decision-making with powerful abstract tools. Such modularity is key for future expansion, enabling the agent to flexibly choose between direct graph access (e.g., via \texttt{SearchGraph}) for novel queries and controlled tools for established needs, enhancing adaptability and reliability.

\subsection{System Design for Real-World Application and AI Advancement}

INRAExplorer demonstrates the effective application of multiple AI techniques to a significant real-world challenge: navigating and reasoning over complex scientific knowledge. Its distinctiveness arises from the synergistic integration of an agentic framework \citep{singh2025agenticsurvey} with deep Knowledge Graph querying capabilities \citep{zhang2025surveygraphrag}. This combination enables sophisticated multi-hop reasoning and the retrieval of exhaustive, structured answers, addressing needs beyond typical RAG systems. The system's architecture leverages key open-source components—including Mirascope for agent orchestration, Qdrant for vector storage, Neo4j for the knowledge graph, GROBID for PDF processing \citep{lopez2009grobid}—and utilizes the open-weight model deepseek-r1-0528. This open design ensures inherent adaptability, meaning INRAExplorer not only serves as a potent tool for its current domain but also offers an extensible architecture. Such a foundation can facilitate the practical integration and evaluation of new AI approaches for advanced information interaction and complex reasoning tasks.

%%%%%%%%%%%%%%%%%%%%%%%%%%%%%%%%%%%%%%%%%%%%%%%%%%%%%%%%%%%%%%%%%%%%%%%%
% Section 4: Conclusion and Future Work
%%%%%%%%%%%%%%%%%%%%%%%%%%%%%%%%%%%%%%%%%%%%%%%%%%%%%%%%%%%%%%%%%%%%%%%%

\section{Conclusion and Future Work}

The INRAExplorer method advances beyond conventional Retrieval-Augmented Generation \citep{lewis2020retrieval} by synergizing an agentic, multi-tool framework \citep{singh2025agenticsurvey} with the deep, structured querying capabilities of Knowledge Graphs (KGs) \citep{zhang2025surveygraphrag}. This distinct combination allows for exhaustive retrieval and sophisticated handling of complex relational queries, offering greater flexibility in problem decomposition and dynamic tool use \citep{yao2023react, schick2023toolformer} than systems focusing narrowly on either general agency or KG querying alone \citep{edge2024graphrag, he2024gretriever}. INRAExplorer thus delivers comprehensive, structured answers crucial for knowledge-intensive domains and represents an open, deployable, and adaptable solution for advanced knowledge exploration, fostering accessibility and community development.

Future work will focus on developing a tailored evaluation framework for INRAExplorer. Standard benchmarks fail to capture the complexity of scientific, multi-hop queries central to our use case. A meaningful assessment requires collaboration with domain experts to define realistic tasks, gold standards, and success criteria. This co-designed approach will enable rigorous validation of both system performance and underlying technical choices in real-world conditions.

Furthermore, looking beyond the initial evaluation, another promising direction for enhancing INRAExplorer involves specializing the core agent model. Techniques inspired by reinforcement learning, such as Reinforcement Learning from Verifiable Feedback (RLVF), could be explored to further refine smaller, more efficient language models. This approach could train them to better navigate the complexities and variable nature of multi-hop reasoning tasks, potentially leading to more robust and nuanced system performance.

%%%%%%%%%%%%%%%%%%%%%%%%%%%%%%%%%%%%%%%%%%%%%%%%%%%%%%%%%%%%%%%%%%%%%%%%
%%% Use this environment to include acknowledgements (optional).
%%% This will be omitted in doubleblind mode.

\begin{ack}
The authors wish to express their gratitude to several individuals and institutions for their contributions to this work. We thank Tristan Salord, Alban Thomas, Eric Cahuzac, François-Xavier Sennesal, Odile Hologne and Hadi Quesneville from INRAE for their assistance in the data gathering process, for providing expert guidance on the rules for merging diverse data sources, for their insightful expert opinions, and for INRAE's role in co-financing this project. We also extend our sincere thanks to Nicolas Chesneau from Ekimetrics for his support and insightful contributions.
\end{ack}

%%%%%%%%%%%%%%%%%%%%%%%%%%%%%%%%%%%%%%%%%%%%%%%%%%%%%%%%%%%%%%%%%%%%%%%%
%%% Use this command to include your bibliography file.

\bibliography{mybibfile}

\end{document}